# Classification of Anuran Frog Species Using Machine Learning

Miriam Alabi

*Abstract*— Acoustic classification of frogs has gotten a lot of attention recently due to its potential applicability in ecological investigations. Numerous studies have been presented for identifying frog species, although the majority of recorded species are thought to be monotypic. The purpose of this study is to demonstrate a method for classifying various frog species using an audio recording. To be more exact, continuous frog recordings are cut into audio snippets first (10 seconds). Then, for each ten-second recording, several time-frequency representations are constructed. Following that, rather than using manually created features, Machine Learning methods are employed to classify the frog species. Data reduction techniques; Principal Component Analysis (PCA) and Independent Component Analysis (ICA) are used to extract the most important features before classification. Finally, to validate our classification accuracy, cross validation and prediction accuracy are used. Experimental results show that PCA extracted features that achieved better classification accuracy both with cross validation and prediction accuracy.

## I. INTRODUCTION

Frogs are a group of small, tailless amphibians with a short body. They are mostly carnivorous and have a lot of different types. They are usually called Anura, which means "without a tail" in Ancient Greek. Frogs can be found all over the world, from the tropics to the subarctic. The most species can be found in tropical rainforests, but there are many more frogs in the subarctic.

Frog populations worldwide have been quickly dropping. Habitat loss, alien species, and climate change are all being blamed for the decline. To study the development of the frog population and enhance its protection strategy, it is necessary to amass information about frogs and their environment. In comparison to traditional approaches, which require ecologists to attend fields often to gather biodiversity data, an acoustic sensor enables the collection of data over a broader range of geographical and temporal scales [2]. Due to the fact that an acoustic sensor may create massive amounts of acoustic data, enabling automated ways to examine acquired data is in great demand. Numerous earlier research have created various strategies for categorizing frog species based on their auditory characteristics [3]. Various feature reduction approaches and classifiers were investigated in those research for frog classification. Mel-Frequency Cepstral Coefficients (MFCCs) are a well-known characteristic for identifying frog sounds [3]. The dataset was utilized in a variety of classification tasks linked to the difficulty of recognizing anuran species based on their sounds.

The Anura dataset was first downloaded from the University of California, Irvine's machine learning repository (https://archive.ics.uci.edu/ml/datasets/Anuran+Calls+%28MFCCs% 29). For this study, certain modifications were done to the data set. The dataset contains 7195 cases for training and testing the 22-attribute classifier model. Also, the dataset was made by splitting up 60 audio files from ten different species and adding a single column (class) of labels. These recordings were made in the same place where they were made and with the same noise level as in the real world (the background sound). One was found in Crdoba, Argentina, and the other species were found in Mata Atlantica, Brazil, and the third was found in Manaus, Brazil. The recordings were made in the WAC4 format, which has a sampling rate of 44.1 kHz and a resolution of 32 bits. This means that we can analyze signals up to 22 kHz.

In this project, we seek to find if machine learning can be a useful tool in classifying frog calls and also to find which feature reduction technique is used in machine learning can reduce the number of features or reduce data dimensions before classification without losing majority of the data.

## II. LITERATURE REVIEW

*Machine Learning Algorithms for Classification*

Machine learning and deep learning are soft computing approaches that originated from artificial intelligence's pattern recognition and computational learning theories [7]. Algorithms for machine learning have been investigated in a wide range of domains and have shown substantial success in a number of them. As evidenced by [8-14], deep learning algorithms have made substantial breakthroughs in the sector of transportation. [15-16] have also effectively used machine learning to the area of industrial engineering. Due to the widespread popularity of machine learning algorithms in a variety of domains, they have been employed for fraud detection in a variety of instances. This computational approach makes use of the research and development of algorithms capable of learning from and forecasting data. In contrast to the job-flow type of program output, these algorithms may construct models using training datasets of observed data and then provide data-driven predictions or judgements as outputs [17]. There are several machine-learning techniques available for constructing regression and classification models. [18] shown how Support Vector Machines (SVMs) may be used to significantly reduce the number of labeled training examples necessary in both inductive and transductive scenarios. [19] automatically classified and mined text or phrases included inside a document using K-nearest neighbor (KNN). [20] categorizes and reduces biases using linear discriminant analysis. [21] examined the use of decision trees (CART) and Naive Bayes (NB) algorithms for detecting and categorizing network applications associated with packet traffic flows during surveillance.

## III. METHODOLOGY

The purpose of this research is to compare the performance of several categorization models created using a variety of machine learning methods. Our technique for classifying frog calls involves four steps: data description, pre-processing, feature extraction, and classification. The next sections detail each stage in detail, while Fig. 1 summarizes them.

### A. Pre-processing

Machine Learning is the process through which Artificial Intelligence is used to allow computers to learn a task from scratch without explicitly programming them. This procedure begins with providing them with high-quality data and continues with training the computers via the construction of various machine learning models utilizing the data and various methods. The method we choose is determined by the kind of data distribution we have and the type of job we are attempting to automate (classification, grouping, or regression). The pre-processing of data is the first stage in the machine learning technique. The phrase "pre-processing" refers to the actions conducted on our data prior to it being sent to the algorithm. Data preparation is a technique that converts unstructured data to structured data. In other words, if data is obtained from several sources, it is collected in an unprocessed format, making analysis impossible. To get the best results from the used model in Machine Learning projects, the data must be properly formatted. This research employs three distinct data pretreatment approaches for machine learning: outlier removal, standardization, and normalization.

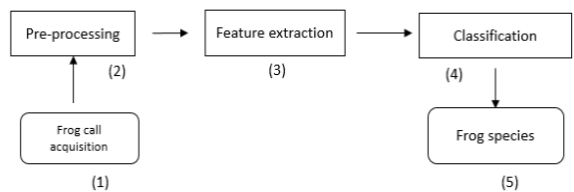

Fig. 1. Flowchart of our frog call classification system

Fortunately, the dataset contains no missing data, and hence we can continue to eliminating outliers using Equation 1. Assuming the distribution of the present dataset is normal, 97 percent of the data values will fall inside the region of a Gaussian curve with limits of -3sigma to +3sigma. That is, any dataset that deviates from these bounds is deemed an outlier.

$$Outlier = -3\sigma \ to \ +3\sigma \quad \quad (1)$$

The dataset is visualized using some descriptive statistics analysis. The data is then transformed by normalizing and standardizing the dataset. This is done to bring all data points to one percentage or narrower range before comparing results.

### B. Features of the data

#### i. Visualizing and relationship between attributes

The correlation matrix is used to depict the statistically linear connection between the characteristics. To create a correlation matrix, Pearson or Spearman Correlations are employed and recorded in a matrix, as illustrated in Figure 2. Additionally, the correlation matrix is utilized to examine the interdependence of numerous qualities at the same time. The output is a table providing the coefficients of correlation between each characteristic and the others. The coefficient yields a number between -1 and 1, indicating the correlation's upper and lower boundaries, from complete negative correlation to complete positive correlation. A correlation coefficient of 0 indicates that there is none. The value must be evaluated, with values less than or equal to -0.5 indicating a significant correlation, and values more than or equal to 0.5 indicating a less significant connection. The matrix's diagonal represents the covariance between each property and itself. The remaining values in the matrix denote the covariance of the two qualities.

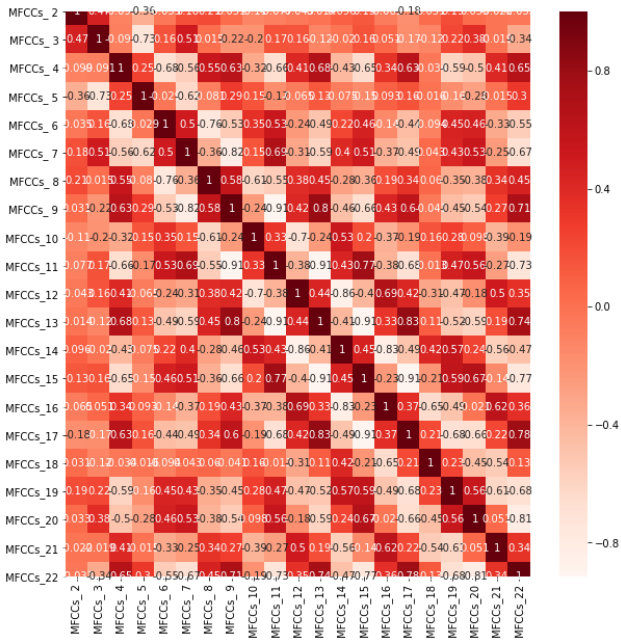

Fig. 2. Heat map of the Correlation matrix

### ii. Feature Extraction

Feature Extraction techniques are then used to reduce the number of features for classification in an attempt to increase performance and reduce the computation time. Only features that account for over 96% of the variance within the data are extracted. The extraction tools used in this study are the principal component analysis (PCA) and independent component analysis (ICA).

- *Principal Component Analysis (PCA)*

PCA generates important features by orthogonally converting correlated data into linearly uncorrelated variables called principle components [5]. The main components analysis combines the majority of the characteristics in order to capture the greatest amount of variation in the data. The dimension of the cleaned data is lowered in this research as a result of the modified characteristics.

- *Independent Component Analysis (ICA)*

ICA decomposes a multivariate signal into additive subcomponents. This is done by assuming that the subcomponents are neither Gaussian nor statistically independent [6].

### iii. Classification Model Development

This study used six classic Supervised Machine Learning (SML) algorithms to build the model and classify the frog species: linear discriminant analysis (LDA), k-nearest neighbor (KNN), logistic regression (LR), classification and regression trees (CART), Naive-Bayes (NB), and support vector machine (SVM) (SVM). These are the most extensively used models due to the ease with which they may be implemented and the great accuracy with which they perform [4]. Python will be used to develop and execute the algorithms in question. When making predictions, the SML algorithms will be used to learn the mapping of historical data, which will then be utilized to construct the model and make predictions.

### iv. Evaluation Metrics

The performance of the models' predictions was assessed using two metrics: cross-validation accuracy (CVA) and prediction accuracy (recall). CVA is a statistical technique that is used to determine the skill of machine learning models. This project makes use of a tenfold cross validation. Recall is a performance metric for classification problems in machine learning, where accuracy may be calculated as in equation 2.

$$Recall = \frac{TP}{(TP + FN)} \ldots \ldots \ldots \ldots \quad (2)$$

Where, True Positive (TP) is when the predicted value is true, False Negative (FN) is when the predicted negative is false

Each data form; raw data, clean data, normalized data, standardized data, PCA, ICA produced 6 classification models for each performance metric.

## IV. RESULTS

To maximize the performance of the models, pre-processing was done on the data. Figure 3 and 4 visualizes the raw data and preprocessed data. It can be seen the outliers have been reduced in Fig. 3.

### A. Data Preprocessing
#### i. Removing outliers

The data begun with 7195 instances and 22 attributes. The cleaning of the data starts with the removal of missing values and outliers. After the outliers have been removed shape of data is reduced to 5416 instances and 22 attributes. A total of 1779 outliers were identified and removed as evaluated by equation (1).

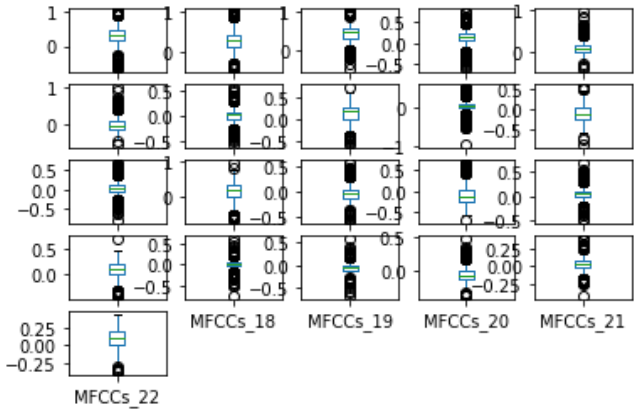

Fig. 3. The boxplot with outliers matrix for the raw dataset.

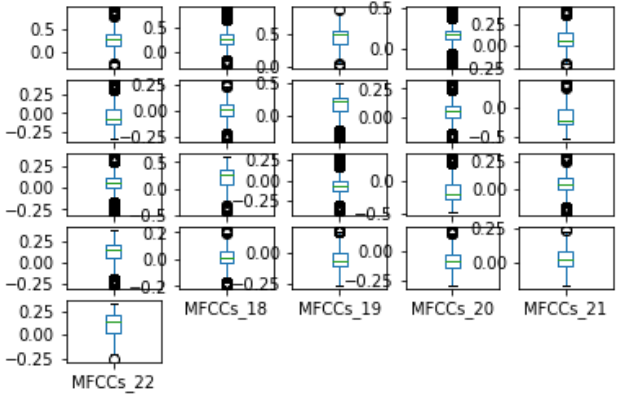

Fig. 4. The boxplot with outliers matrix for the dataset after the removal of outliers.

### ii. Normalization and standardization

After the removal of the outliers, the data is normalized to bring all data points between 0 and 1. Standardization is also done to ensure that all the data points is converted into a z score value, where they follow the Gaussian curve as shown in Fig.5.

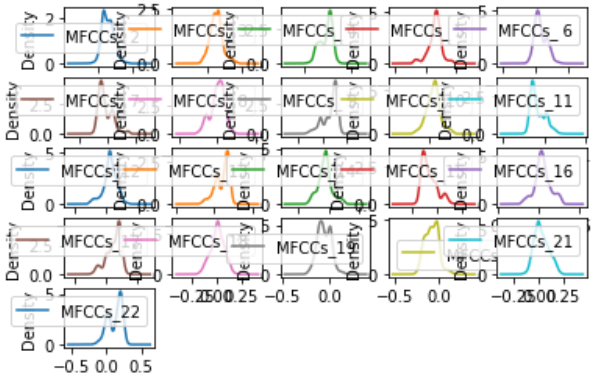

Fig. 5. The density plots of the standardized dataset showing a normal distribution

### B. Feature extraction techniques

After preprocessing, there may be complex and unknown relationships between the attributes in the dataset. To quantify and discover the degree to which the attributes in the dataset are dependent on each other, the correlation matrix was used. In Figure 2, it is seen that attribute MFCCs 13 and MFCCs 17 are highly positive correlated with a correlation coefficient of 0.83. This means that attributes of high positive correlation depend on each other (both change in the same direction) or in this case 83 % of MFCCs 13 contents can be explained by MFCCs 17. There were quite a number of attributes with a low positive correlation of 0.01. For example between MFCCs 5 and 18, MFCCs 5 and 2. This shows that they were not correlated in any way and do not depend on each other. MFCCs 9 and 11, MFCCs 11 and 13, MFCCs 13 and 15, MFCCs 15 and 17, show very high negative correlation coefficient of -0.91. This explains that all the attributes change in opposite directions. This knowledge helps in preparation of the dataset to meet the expectations of the machine learning algorithms. In classification, performance will degrade with the presence of these interdependencies.

This step attempted to eliminate redundancy in the data by converting it to capture the bulk of the variation while lowering the dimensions, or by simply picking the important qualities that account for the most of the variance. Following PCA and ICA, the data was converted into 5416 instances and ten characteristics. For PCA, PCA explained variance ratios: [0.47852, 0.168779, 0.093008, 0.070274, 0.043009, 0.03567, 0.026017, 0.01991, 0.015739, 0.008655] and sum of PCA ratios is 0.9596 indicating that 10 attributes were able to explain approximately 96% of the variance within the data after. For ICA, the technique extracted 10 attributes after 21 iterations.

### C. Machine Learning classification algorithms and evaluation

#### i. Cross-validation Accuracy (CVA) train

In this section only 80% of the data is used to train the model. In Table 1 the cross validation accuracy is shown for all data types.

*Table1:* Performance of models based of cross-validation Accuracy.

| Data | LR | LDA | KNN | CART | NB | SVM | Mean |
|---|---|---|---|---|---|---|---|
| Raw | 0.9270 | 0.9408 | 0.9821 | 0.9392 | 0.9343 | 0.9222 | 0.9409 |
| Clean | 0.9414 | 0.9744 | 0.9954 | 0.9651 | 0.9691 | 0.9248 | 0.9617 |
| Norm | 0.9573 | 0.9744 | 0.9939 | 0.9624 | 0.9691 | 0.9549 | 0.9687 |
| Stand | 0.9792 | 0.9744 | 0.9947 | 0.9654 | 0.9691 | 0.9963 | **0.9799** |
| PCA | 0.9287 | 0.9693 | 0.9933 | 0.9686 | 0.9660 | 0.9448 | 0.9618 |
| ICA | 0.6034 | 0.9693 | 0.9864 | 0.9519 | 0.9474 | 0.6034 | 0.8436 |

The results are consistent within the values but the ICA

transformation, produced very low mean accuracy compared to the mean of the raw data mainly in the SVM model. The normalized, standardized and PCA transformation produced higher mean accuracy than the raw data type. The transformation that produced the highest accuracy is the standardization of the data. KNN model fairly showed good accuracy throughout the data types.

### ii. Outcome of prediction accuracy (test)

In this part, only 20% of the data is used to test the model. Recall is used to see if 20 percent of the data could correctly predict the model, which is how it shows how well it works. The results are shown in this text is summarized in Table2. The results show some level of consistency within the values but the ICA transformation produced very low mean prediction accuracy mainly in the SVM model. The normalized, standardized and PCA transformation produced higher mean accuracy than the raw data type. The transformation that produced the highest accuracy is the standardization of the data. KNN model fairly showed good accuracy throughout the data types.

TABLE 2: Performance of models based of Prediction Accuracy.

| Data | LR | LDA | KNN | CART | NB | SVM | Mean |
|---|---|---|---|---|---|---|---|
| Raw | 0.9430 | 0.9507 | 0.9889 | 0.9507 | 0.9367 | 0.9375 | 0.95125 |
| Clean | 0.9409 | 0.9779 | 0.9917 | 0.9705 | 0.9732 | 0.9373 | 0.96525 |
| Norm | 0.9585 | 0.9779 | 0.9889 | 0.9677 | 0.9732 | 0.9613 | 0.97125 |
| Stand | 0.9806 | 0.9779 | 0.9880 | 0.9714 | 0.9732 | 0.9945 | 0.980933 |
| PCA | 0.9289 | 0.9677 | 0.9899 | 0.9723 | 0.9667 | 0.9493 | 0.962467 |
| ICA | 0.6125 | 0.9677 | 0.9852 | 0.9659 | 0.9493 | 0.6125 | 0.84885 |

## V. DISCUSSION AND CONCLUSIONS

Machine-learning (ML) algorithms offer a unique advantage in developing classification models for categorizing Anura Frog species. These models do not require any prior assumptions to be satisfied before their implementation. However, insufficient or noisy dataset can significantly affect the performance of these algorithms. KNN was seen to outperform the other algorithms after transformation. During the ICA transformation, the algorithms did not show high values of accuracy, hence in an extension to this project only PCA will be used as a feature extraction technique. In this project, we employed six classification algorithms and assessed their performance when creating models, as well as predict with the models.

With the computational analyses, many of the individual algorithms produced adequate classification accuracy, and performance both in the training and testing stage.

ACKNOWLEDGMENT

The author would like to thank Dr. Balakrishna Gokaraju for tutoring and giving an insightful approach to Machine learning.